\documentclass{article}
\usepackage{spconf,amsmath,graphicx}
\usepackage{times}
\usepackage{epsfig}
\usepackage{graphicx}
\usepackage{amsmath}
\usepackage{amssymb}
\usepackage{booktabs}
\usepackage{float}
\usepackage{algorithm}
\usepackage{algorithmic}
\usepackage{subfigure}

\title{Diffusion Motion: Generate Text-Guided 3D Human Motion by Diffusion Model}
\name{Zhiyuan Ren, Zhihong Pan, Xin Zhou, Kang Le}
\address{Michigan State University\\
Baidu Research (USA),  Sunnyvale, CA, 94089, USA}

\twoauthors
 {Zhiyuan Ren}
	{Michigan State University\\
	East Lansing, MI 48824, USA}
 {Zhihong Pan, Xin Zhou, Le Kang}
	{Baidu Research (USA)\\
	Sunnyvale, CA, 94089, USA}
	
\begin{document}
%
\maketitle
%
\begin{abstract}
    We propose a simple and novel method for generating 3D human motion from complex natural language sentences, which describe different velocity, direction and composition of all kinds of actions. Different from existing methods that use classical generative architecture, we apply the Denoising Diffusion Probabilistic Model to this task, synthesizing diverse motion results under the guidance of texts. The diffusion model converts white noise into structured 3D motion by a Markov process with a series of denoising steps and is efficiently trained by optimizing a variational lower bound. To achieve the goal of text-conditioned image synthesis, we use the classifier-free guidance strategy to add text embedding into the model during training.  Our experiments demonstrate that our model achieves competitive results on HumanML3D test set quantitatively and can generate more visually natural and diverse examples. We also show with experiments that our model is capable of zero-shot generation of motions for unseen text guidance. 
\end{abstract}

\section{Introduction}

Generating 3D human motion from natural language sentences is an interesting and useful task. It has extensive  applications across virtual avatar controlling, robot motion planning,  virtual assistants and movie script visualization. 

The task has two major challenges. First, since natural language can have very fine-grained representation, generating visually natural and semantically relevant motions from texts is difficult. Specifically, the text inputs can contain a lot of subtleties. For instance, given different verbs and adverbs in the text, the model needs to generate different 
motions. The input may indicate different velocities or directions, e.g., ``a person is running fast forward then walking slowly backward". The input may also describe a diverse set of motions, e.g., ``a man is playing golf", ``a person is playing the violin ", ``a person walks steadily along a path while holding onto rails to keep balance". The second challenge is that one textual description could 
map to multiple motions. This requires the generative model to be probabilistic. For instance, the generated motions from the description ``a person is walking" should have multiple output samples with different velocities and directions.

Early motion generating methods \cite{guo2020action2motion, cai2018deep, tulyakov2018mocogan, petrovich2021action} in generating 3D human motions are based on very simple textual descriptions, such as an action category, e.g. jump, throw or run. This type of setup has two limitations. First, the feature space of input texts is too sparse. Therefore, the solutions do not generalize to texts outside the distribution of the dataset. Second, category-based texts have very limited applications in real-world scenarios. With the emergence of the KIT-ML dataset \cite{plappert2016kit}, which contains 6,278 long sentence descriptions and 3,911 complex motions, a series of work \cite{ahuja2019language2pose, ghosh2021synthesis} started to convert complex sentence modalities into motion modalities. They usually design a sequence-to-sequence architecture to generate one result. However, this is inconsistent with the nature of the motion generation task because every language modality corresponds to a very diverse set of 3D motions. Most recently, a new dataset HumanML3D and a new model has been proposed in \cite{guo2022generating} to solve the above problems. The dataset consists of 14,616 motion clips and 44,970 text descriptions and provides the basis for training models that can generate multiple results. The new model proposed in \cite{guo2022generating} is able to generate high-fidelity and multiple samples. It achieves state-of-the-art quantitatively. However, the generated samples have very limited diversity and are not capable of achieving zero-shot test. In addition, this model consists of several sub-models, which cannot be trained end-to-end, and the inference process is very complex.

A new paradigm for image and video generation named denoising diffusion probabilistic models, has recently emerged and achieved remarkable results \cite{ho2020denoising, song2020denoising}. The diffusion model learns an iterative denoising process that gradually recovers the target output from a Gaussian noise at inference time. Many recent papers aim to generate images based on textual descriptions. They blend text into the input and guide the generation of images using techniques such as classifier guidance \cite{dhariwal2021diffusion}, classifier-free guidance \cite{ho2022classifier}, which synthesize impressive samples, such as \cite{ramesh2022hierarchical}. Generation using diffusion models are also applied to other modalities such as speech generation \cite{kong2020diffwave} and point cloud generation \cite{luo2021diffusion} to achieve much better results than previously possible.

We apply the diffusion model to the task of 3D human motion generation based on textual descriptions. Our results show that the generated samples have better fidelity and wider diversity. The guidance of texts is more controllable. More specifically, we make the following contributions:

\begin{itemize}
    \item To our best knowledge, we are the first to construct the diffusion motion architecture for 3D human motion generation based on textual description.
    \item Based on visualized experimental results, we discuss the advantages of our diffusion model in terms of flexibility in text control, diversity of generated samples, and zero-shot capability for motion generation.
    \item Quantitatively, we have achieved very competitive  results in our experiments with existing metrics on the HumanML3D test set.
\end{itemize}


\section{Method}
\begin{figure}[h]
    \centering
    \includegraphics[width=0.5\textwidth]{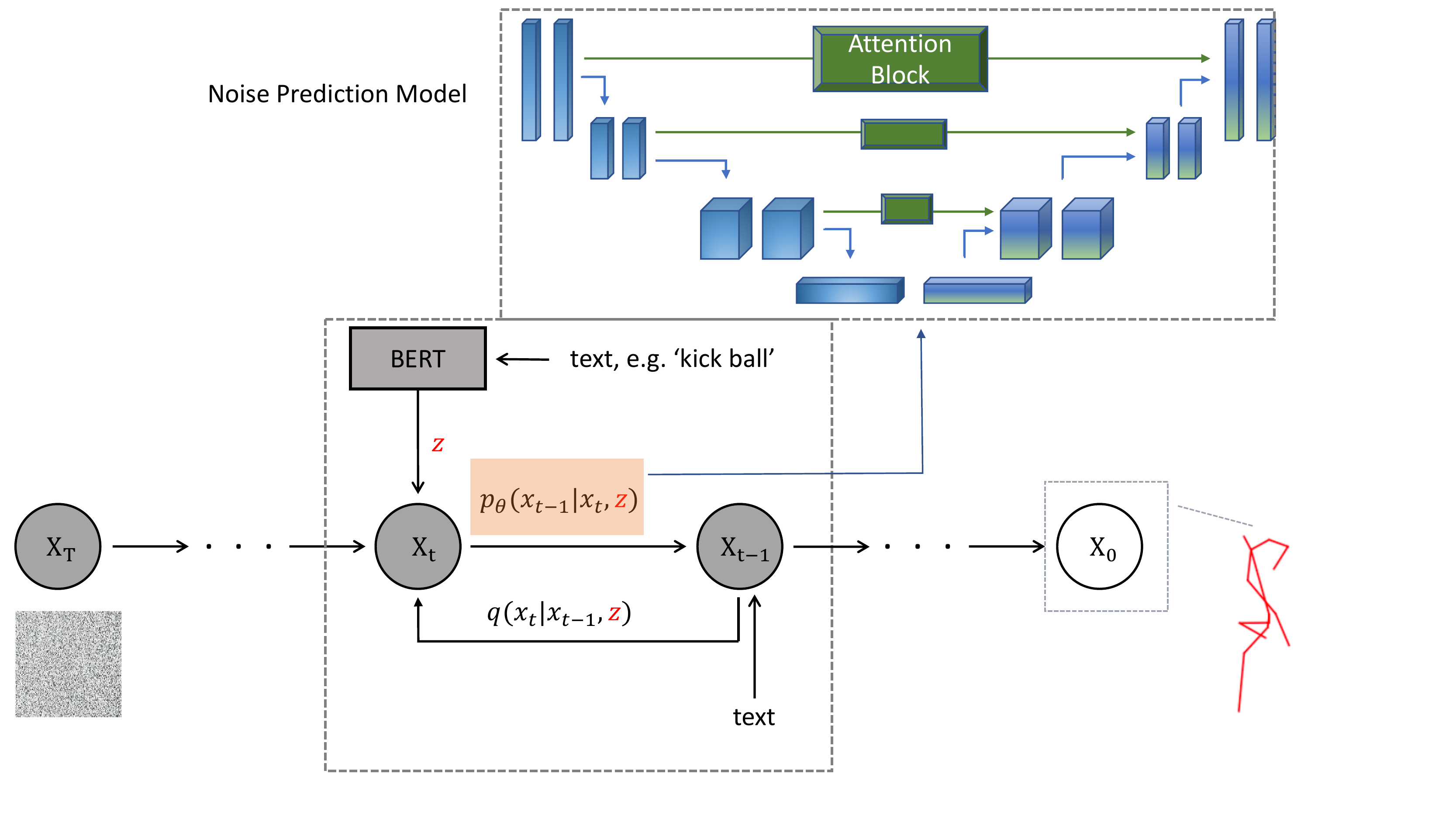}
    \caption{Architecture overview. The forward process is a Markov process without training parameters; The reverse process uses a Unet-like architecture to predict the noise in every step. We utilize a pre-trained BERT model for feature extraction for text input.}
    \label{fig:1}
    \vspace{-5mm}
\end{figure}

In this section, we first formulate the probabilistic model of forward and reverse diffusion processes for 3D human motion generation from text descriptions. Then, we explain the mathematical expression of the modified objective for training the model. Lastly, we write the full training and sampling algorithm. The whole architecture is shown in Fig \ref{fig:1}.

\subsection{Diffusion Processes}
Each 3D human pose is represented by the position of $J$  keypoints. Each keypoint has three coordinates in 3D. Therefore, a 3D human pose can be denoted by $\mathbf{p}  \in \mathbb{R}^{J \times 3}$. 3D human motion is a sequence of poses and hence can be written as $\mathbf{x}  \in \mathbb{R}^{L\times J \times 3}$ where $L$ represents the number of time steps. For simplicity, we flatten the pose representation into a one-dimensional vector. Then the motion can be represented as $\mathbf{x}  \in \mathbb{R}^{L\times C}$ where $C = J \times 3$. 

The denoising diffusion probabilistic model consists of a forward process and a reverse process. The forward diffusion process converts the original highly structured and semantically relevant keypoints distribution into a Gaussian noise distribution. We denote motion with increasing levels of noise as $\mathbf{{x}_{0}}, \mathbf{{x}_{1}}, ...,\mathbf{{x}_{T}}$. The forward diffusion process is modeled as a Markov chain:
\begin{equation}
q\left(\mathbf{x}_{1: T} \mid \mathbf{x}_{0}\right)=\prod_{t=1}^{T} q\left(\mathbf{x}_{t} \mid \mathbf{x}_{t-1}\right)
\end{equation}
where $q\left(\mathbf{x}_{t} \mid \mathbf{x}_{t-1}\right)$ is denoted as the forward process, adding noise to motions at the previous time step and generate the distribution at the next time step. Given a set of pre-defined hyper-parameters $\beta_{1}, \beta_{2},...,\beta_{T}$, each transition step can be written as 
\begin{equation}
q\left(\mathbf{x}_{t} \mid \mathbf{x}_{t-1}\right):=\mathcal{N}\left(\sqrt{1-\beta_{t}} \mathbf{x}_{t-1}, \beta_{t} \mathbf{I}\right)
\end{equation}
where $\beta_{t}$ controls the diffusion rate of the process and usually ranges from 0 to 1.

The reverse diffusion process is a generation process where keypoints sampled from a Gaussian noise distribution $p\left(\mathbf{x}_{t}\right)$ are given as the initial input. Then the initial input is progressively denoised on an inverse Markov chain following the given text description. The text description is encoded into a latent variable $\mathbf{z}$ using the BERT \cite{devlin2018bert} model. $\theta$ are the parameters of the denoising model. The reverse diffusion process can be written as follows:
\begin{equation}
\resizebox{.8\hsize}{!}{$p_{\boldsymbol{\theta}}\left(\mathbf{x}_{0: T} \mid \mathbf{z}\right)=p\left(\mathbf{x}_{T}\right) \prod_{t=1}^{T} p_{\boldsymbol{\theta}}\left(\mathbf{x}_{t-1} \mid \mathbf{x}_{t}, \mathbf{z}\right)$}
\end{equation}
\vspace{-2mm}
\begin{equation}
\resizebox{.9\hsize}{!}{$p_{\boldsymbol{\theta}}\left(\mathbf{x}_{t-1} \mid \mathbf{x}_{t}, \mathbf{z}\right)=\mathcal{N}\left(\mathbf{x}_{t-1} \mid \boldsymbol{\mu}_{\boldsymbol{\theta}}\left(\mathbf{x}_{t}, \mathbf{t}, \mathbf{z}\right), \beta_{t} \boldsymbol{I}\right)$}
\end{equation}
where $\boldsymbol{\mu}_{\boldsymbol{\theta}}$ is the target we want to estimate by a neural network. $t$ is the timestep indicating where the denoising process has conducted, which is encoded as a vector based on the cosine schedule \cite{nichol2021improved}. $p\left(\mathbf{x}_{t-1} \mid \mathbf{x}_{t}, \mathbf{z}\right)$ represents the reverse transitional probability of keypoints from one step to the previous step.

\subsection{Training Objective}
In the previous subsection, we established the forward and reverse diffusion process for 3D human motion generation from textual descriptions. In this subsection, we will derive the mathematical expressions for our training objective of the reverse diffusion process. 

In order to approximate the intractable marginal likelihood $p_{\boldsymbol{\theta}}\left(\mathbf{x} \right)$, we maximize a variational lower bound to make the objective optimal:
\begin{equation}
\mathrm{E}_{q\left(\mathbf{x}_{0}\right)}\left[\log p_{\theta}\left(\mathbf{x}_{0}\right)\right] \geq \mathrm{E}_{q\left(\mathbf{x}_{0: T}\right)}\left[\log \frac{p_{\theta}\left(\mathbf{x}_{0: T}, \mathbf{z}\right)}{q\left(\mathbf{x}_{1: T},  \mathbf{z}\mid \mathbf{x}_{0}\right)}\right]
\end{equation}
It is the same objective as in DDPM \cite{ho2020denoising} except for involving the text embedding $\mathbf{z}$. We add text embedding $\mathbf{z}$ and time embedding $\mathbf{t}$ into a new embedding during our specific training process. For simplicity, we still represent the fused embedding with $\mathbf{t}$, which makes the objective consistent with DDPM:
\begin{equation}
\mathrm{E}_{q\left(\mathbf{x}_{0}\right)}\left[\log p_{\theta}\left(\mathbf{x}_{0}\right)\right] \geq \mathrm{E}_{q\left(\mathbf{x}_{0: T}\right)}\left[\log \frac{p_{\theta}\left(\mathbf{x}_{0: T}, \right)}{q\left(\mathbf{x}_{1: T}, \mid \mathbf{x}_{0}\right)}\right]
\end{equation}
The same as DDPM simplifies the objective, our final training objective is
\begin{equation}
\left\|\boldsymbol{\epsilon}-\boldsymbol{\epsilon}_{\theta}\left(\mathbf{x}_\mathbf{t}, \mathbf{t}\right)\right\|^{2}, \quad \epsilon \sim \mathcal{N}(0, \mathbf{I})
\end{equation}
where $\epsilon$ is the noise sampled from standard Gaussian distribution, $\boldsymbol{\epsilon}_{\theta}\left(\mathbf{x}_\mathbf{t}, t\right)$ is the output of the noise prediction model.

\subsection{Training Algorithm}
In principle, the training target is to minimize Eq. (7). The simplified training and sampling algorithm is as follows:
\begin{algorithm}
    \caption{Training Process(Simplified)}
    \begin{algorithmic}
        \STATE \textbf{repeat}
        \STATE \hspace{0.5cm}$\mathbf{x}_{0}, \mathbf{z} \sim q\left(\mathbf{x}_{0}\right)$
        \STATE \hspace{0.5cm}$\mathbf{z} \leftarrow \varnothing$ with probability $p_{\text {uncond }}$
        \STATE \hspace{0.5cm}$\mathbf{t} \sim \operatorname{Uniform}(\{1, \ldots, T\})$
        \STATE \hspace{0.5cm}$\lambda \sim p(\lambda)$
        \STATE \hspace{0.5cm}$\boldsymbol{\epsilon} \sim \mathcal{N}(\mathbf{0}, \mathbf{I})$ \STATE \hspace{0.5cm}$\mathbf{x}=\alpha_{\lambda} \mathbf{x}+\sigma_{\lambda} \boldsymbol{\epsilon}$
        \STATE \hspace{0.5cm}$\mathbf{t} = f(\mathbf{t}, \mathbf{z})$
        \STATE \hspace{0.5cm}Take gradient descent step on
        \STATE \hspace{1cm} $\nabla_{\theta}\left\|\boldsymbol{\epsilon}-\boldsymbol{\epsilon}_{\theta}\left(\sqrt{\bar{\alpha}_{t}} \mathbf{x}_{0}+\sqrt{1-\bar{\alpha}_{t}} \boldsymbol{\epsilon}, \mathbf{t}\right)\right\|^{2}$
        \STATE \textbf{until} converged
    \end{algorithmic}
\end{algorithm}
\begin{algorithm}
    \caption{Sampling Process(Simplified)}
    \begin{algorithmic}
        \STATE $\mathbf{x}_{T} \sim \mathcal{N}(\mathbf{0}, \mathbf{I})$
        \STATE $\text { for } t = T, \ldots, 1 \hspace{2mm} \mathbf{do:} $
        \STATE \hspace{0.5cm}$\quad \tilde{\boldsymbol{\epsilon}}_{t}=(1+w) \boldsymbol{\epsilon}_{\theta}\left(\mathbf{x}_{t}, \mathbf{z}\right)-w \boldsymbol{\epsilon}_{\theta}\left(\mathbf{x}_{t}\right)$
        \STATE \hspace{0.5cm}$\quad \triangleright \text { Sampling step }(\text {we adopt the way DDPM does})$
        \STATE \textbf{end for }
        \STATE return $\mathbf{x}_{0}$
    \end{algorithmic}
\end{algorithm}

\section{Experiments}

We present our experimental setup in Section \textbf{3.1}. The quantitative and visualized results are compared to the current SOTA method Temporal VAE \cite{guo2022generating} in Section \textbf{3.2}. In Section \textbf{3.3}, we make some further comments on the evaluation and discuss future work.

\subsection{Experimental Setup}

\noindent \textbf{Datasets.} For 3D human motion generation from textual descriptions, we use the HumanML3D \cite{guo2022generating} dataset originating from a combination of HummanAct12 \cite{guo2020action2motion} and Amass \cite{mahmood2019amass} datasets. It covers a broad range of human actions such as daily activities.

\noindent \textbf{Baseline Methods.} We compare our work to Seq2Seq \cite{linvigil18}, Language2Pose \cite{ahuja2019language2pose}, Text2Gesture \cite{bhattacharya2021text2gestures}, MoCoGAN \cite{tulyakov2018mocogan}, Dance2Music \cite{tang2018dance} and Temporal VAE \cite{guo2022generating}. Except Temporal VAE, all these existing methods either are deterministic methods, which means they can generate only one result from each text input, or are not able to do the 3D human motion generation from textual description task directly. 

\noindent \textbf{Evaluation Metrics.} Our quantitative evaluation is to measure the naturality, semantic relevance and diversity of the generated 3D human motions. Below are details of the four metrics:

\begin{enumerate}
  \item \textbf{Recognition Precision.} We use a pre-trained Motion encoder and Text encoder to compute the generated samples' embedding and text embedding, and further compare the similarity of those.
  \item \textbf{Frechet Inception Distance (FID).} Unlike agreeing to use the inception network for feature extraction for image FID, there is no agreed network for 3D human motion. Meanwhile, to better compare with Temporal VAE's experimental results, we use the same encoder as in their work to measure FID. Their model is both trained and evaluated on this encoder so it puts our method at a disadvantage. We only show this as a reference point.
  \item \textbf{Diversity.} We propose this new metric which evaluates diversity without an encoder. Specifically, diversity measures how much the generated motions diversify with the same text input. Given a set of generated 3D motions with $T$ text inputs. For $t$-th motion, we randomly sample two subsets with the same size $S$, and we use $m$ to represent the generated motion. The diversity can be formalized as:
\begin{equation}
    \text { Diversity }=\frac{1}{T \times S} \sum_{t=1}^{T} \sum_{i=1}^{S}\left\|\mathbf{m}_{t, i}-\mathbf{m}_{t, i}^{\prime}\right\|_{2}
\end{equation}
  
 \begin{table*}[ht!]
    \vspace{-4.5em}%
  \centering
  \begin{tabular}{|c c c c c c c|}
  \hline
  \textbf{Method} & \textbf{R Precision(T-1)$\uparrow$} & \textbf{R Precision(T-2)$\uparrow$} & \textbf{R Precision(T-3)$\uparrow$} & \textbf{FID$\downarrow$} & \textbf{Diversity$\uparrow$} & \textbf{Variance$\rightarrow$} \\ [0.5ex] 
  \hline\hline
  \textbf{Real motions} & 0.511 & 0.703 & 0.707 & 0.002 & 0 & 9.503\\
  \hline
  Seq2Seq \cite{linvigil18} & 0.180 & 0.300 & 0.396 & 11.75 & 0 & 6.223 \\
  Language2Pose \cite{ahuja2019language2pose} & 0.246 & 0.387 & 0.486 & 11.02 & 0 & 7.626 \\
  Text2Gesture \cite{bhattacharya2021text2gestures} & 0.165 & 0.267 & 0.345 & 7.664 & 0 & 6.409 \\
  MoCoGAN \cite{tulyakov2018mocogan} & 0.037 &0.072 & 0.106 & 94.41 & 9.421 & 0.462 \\
  Dance2Music \cite{tang2018dance} & 0.033 & 0.065 & 0.097 & 66.98 & 7.235 & 0.725\\
  \textbf{Temporal VAE} \cite{guo2022generating} & \textbf{0.457} & \textbf{0.639} & \textbf{0.740} & \textbf{1.067} & 18.529 & \textbf{9.188} \\
  \hline
  \textbf{Ours} & 0.406 & 0.612 & 0.735 & 10.21 & \textbf{23.692} & 7.660\\
    \hline
 
  \end{tabular}
  \caption{Quantitative evaluation results on the HumanML3D test set. R Precision(Top-$k$) is short for Recognition Precision with Top-$k$ accuracy. $\uparrow$ means the larger value the better performance; $\downarrow$ means the smaller value the better performance;$\rightarrow$ means the closer to Real motions the better performance.}
  \label{tab:1}
\end{table*}
  \item \textbf{Variance.} This metric measures the variance of the generated samples without considering the text input. For any two subsets of the dataset with the same size $S$, suppose after encoding with the Temporal VAE method, their motion feature vectors are $\left\{\mathbf{v}_{1}, \ldots, \mathbf{v}_{S}\right \}$ and $\left\{\mathbf{v}_{1}^{\prime}, \ldots, \mathbf{v}_{S}^{\prime}\right \}$. The variance of motions is calculated as:
  \begin{equation}
\text { Variance }=\frac{1}{S} \sum_{i=1}^{S}\left\|\mathbf{v}_{i}-\mathbf{v}_{i}^{\prime}\right\|_{2}
\end{equation}
\end{enumerate} 
\newpage

\begin{figure}[h]
    \centering
    \includegraphics[width=0.5\textwidth]{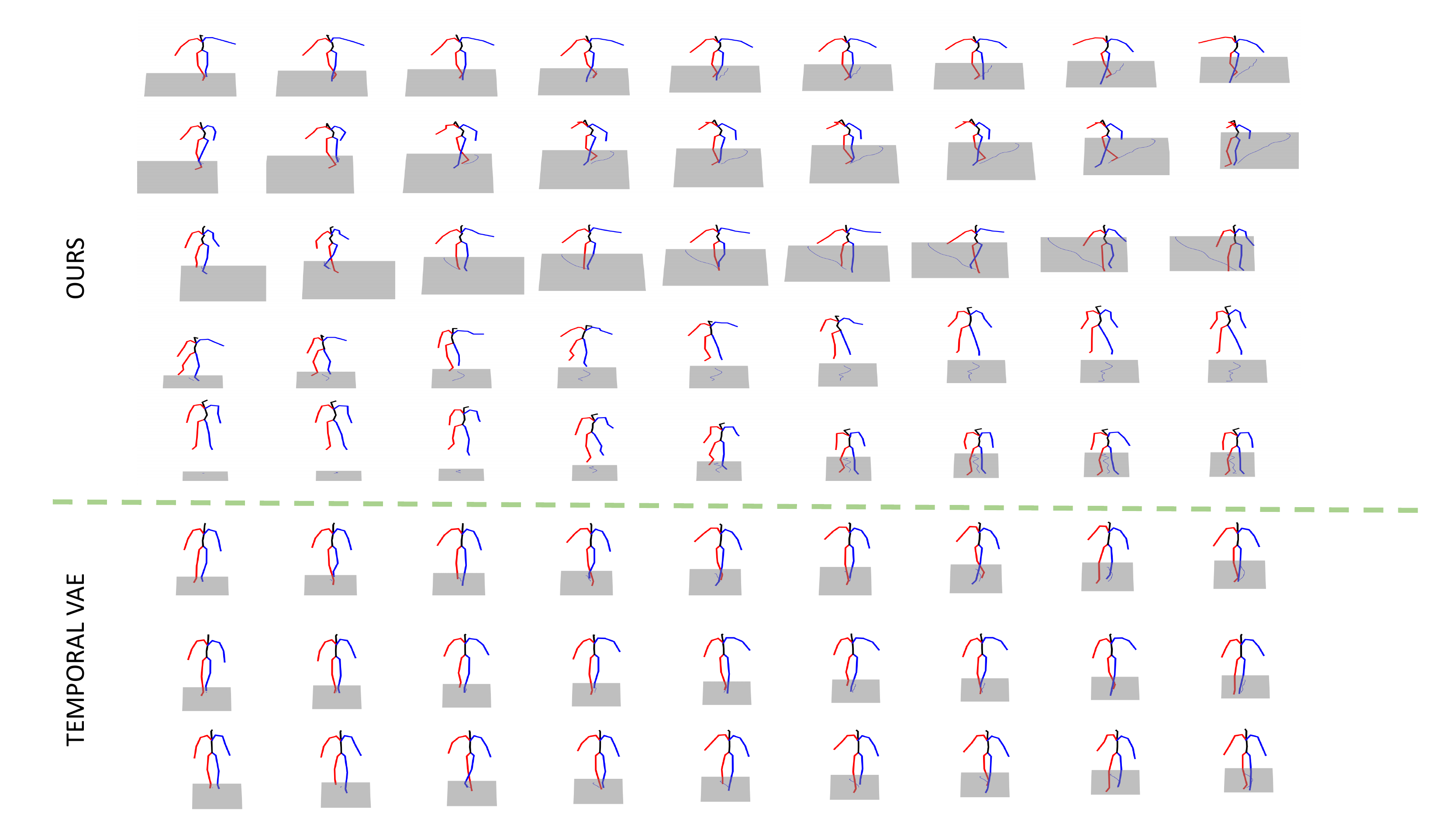}
    \title{A person \textbf{walks steadily along a path} while\textbf{ holding onto rails} to keep balance.}
    \caption{\textbf{Visualized results} of our method vs. those of \cite{guo2022generating}. Given the same text input as the same as the title, we show two generated motions from our method and Temporal VAE. Obviously, for the same output we have much better diversity for the generated samples than Temporal VAE.}
    \label{fig:2}
    \vspace{-5mm}
\end{figure}


\subsection{Quantitative and Visualized Evaluation}

Table \ref{tab:1} shows the quantitative evaluation results for the HumanML3D dataset. For Recognition Precision, our method and Temporal VAE are far more superior than other methods. This indicates that the generated keypoint sequences are highly correlated with the textual descriptions. For the FID and Diversity metrics, Seq2Seq \cite{linvigil18}, Language2Pose \cite{ahuja2019language2pose} and Text2Gesture \cite{bhattacharya2021text2gestures} can generate natural results, but due to their deterministic model design, they have no diversity for this task. MoCoGAN \cite{tulyakov2018mocogan} and Dance2Music \cite{tang2018dance} have unfavorable FID scores but achieve diversity. Clearly, there is a trade-off between the naturality and diversity of the generated results. Overall, Our method and Temporal VAE are far better than other methods in terms of naturality and diversity.

Fig \ref{fig:2} illustrates the visualized results of our method vs. Temporal VAE. For the input ``A person walks steadily along a path while holding onto rails to keep balance'', our method generates motions with different directions, velocities and distances. This shows that the diversity of our method is beyond numerical variances, reaching semantic levels. We do not observe this level of diversity in other models. For instance, the generated results from Temporal VAE contain only minor differences. Our model even interprets `path' to be  `stairs', which is shown in row 4 and row 5. 

\begin{figure}[h]
    \centering
    \title{Zero-shot Test} 
    \includegraphics[width=0.5\textwidth]{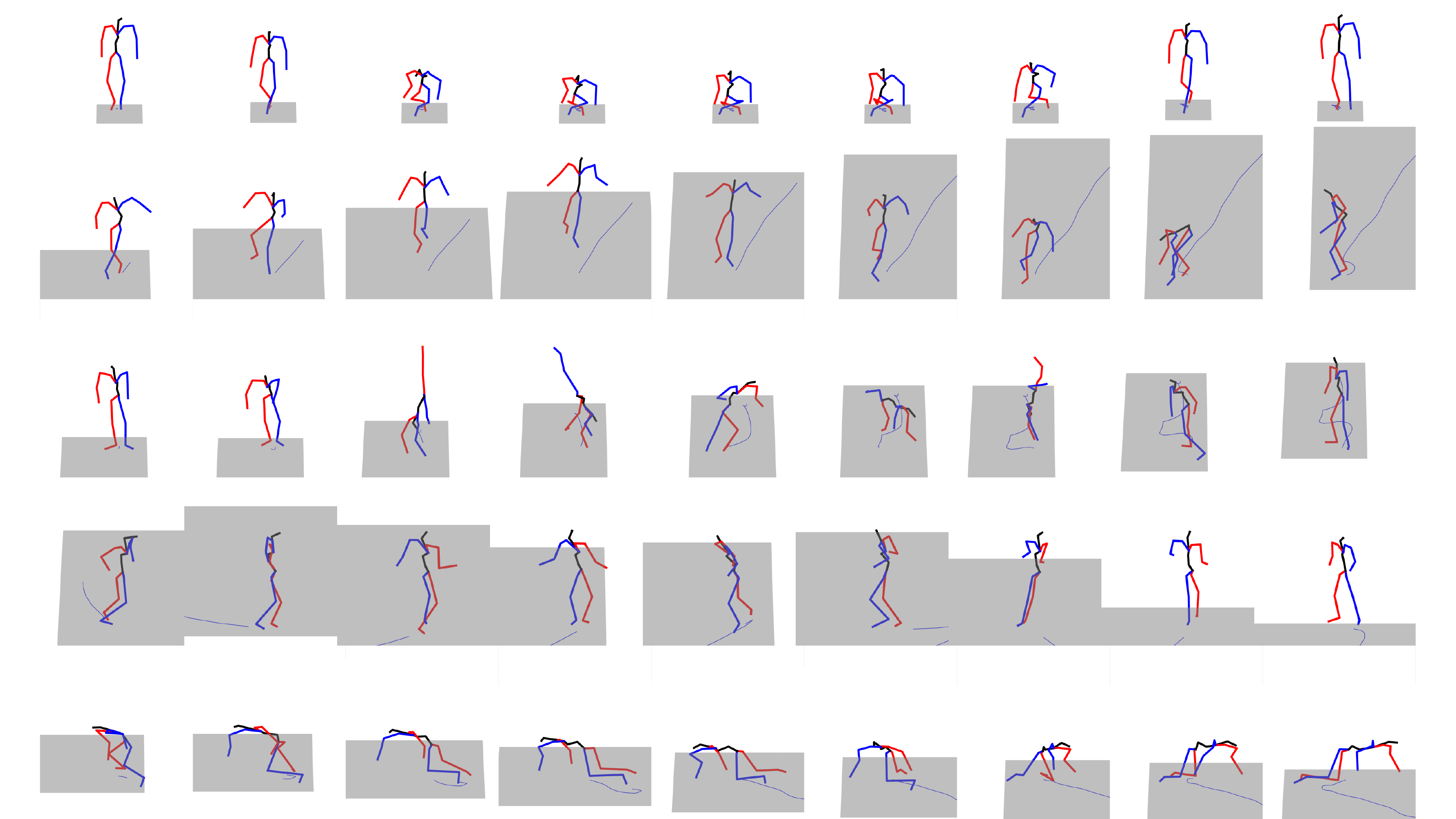}
    \title{A person \textbf{walks steadily along a path} while\textbf{ holding onto rails} to keep balance.}
    \caption{zero-shot text description inputs are below:\\
    row1: a person sits cross-legged for a while, and then suddenly stands up.\\
    row2: a person is jumping hard over a big rock.\\
    row3: a person flips twice in a row.\\
    row4: a person first runs counterclockwise, then walks clockwise, then runs counterclockwise.\\
    row5: a man crawling on the ground with his knees.}
    \label{fig:3}
    \vspace{-5mm}
\end{figure}

Thanks to the pre-trained model BERT \cite{devlin2018bert} and classifier-free guidance technique \cite{ho2022classifier}, our diffusion model has very impressive zero-shot capability. As shown in Fig \ref{fig:3}, for unseen actions and combinations, our method can generate natural and semantically relevant results.


\section{Conclusion}

Our work applies diffusion models to the task of 3D human motion generation based on textual descriptions. We invent three new techniques. We modify the architecture of the denoising model to take 3D motion input. We fuse the BERT embedding into time embedding. We apply the classifier-free guidance technique in this setting. With the largest language-motion dataset HumanML3D, our method performs competitively in quantitative evaluations. Studying the visualized results, we can see that we excel in diversity by a far margin in meaningful ways. We also demonstrate that our method is capable of impressive performance in  zero-shot settings.

\bibliographystyle{IEEEbib}
\bibliography{strings,refs}
\end{document}